\documentclass[conference]{IEEEtran}
\usepackage{cite}
\usepackage{amsmath,amssymb,amsfonts}
\usepackage{algorithmic}
\usepackage{adjustbox}
\usepackage{booktabs}
\usepackage{multirow}
\usepackage{pifont}
\usepackage{graphicx}
\usepackage{textcomp}
\graphicspath{{images/}}
\usepackage{xcolor}
\usepackage{hyperref}
\def\BibTeX{{\rm B\kern-.05em{\sc i\kern-.025em b}\kern-.08em
    T\kern-.1667em\lower.7ex\hbox{E}\kern-.125emX}}
\begin{document}

\title{Masked Multi-Query Slot Attention for Unsupervised Object Discovery
\thanks{This research was supported by the National Science and
Engineering Research Council of Canada (NSERC), via its
Discovery Grant program and MITACS through the Globalink program. The authors acknowledge the support
provided in part by Calcul Quebéc and the Digital Research Alliance of Canada.}
}

\author{\IEEEauthorblockN{Rishav Pramanik}
\IEEEauthorblockA{\textit{Jadavpur University}\\
Kolkata, India \\
rishavpramanik@gmail.com}
\and
\IEEEauthorblockN{José-Fabian Villa-Vásquez}
\IEEEauthorblockA{\textit{ÉTS Montréal}\\
Montréal, Canada \\
jose.villa-vasquez.1@ens.etsmtl.ca}
\and
\IEEEauthorblockN{Marco Pedersoli}
\IEEEauthorblockA{\textit{ÉTS Montréal}\\
Montréal, Canada \\
marco.pedersoli@etsmtl.ca}}

\maketitle
\pagestyle{plain}
\begin{abstract}
Unsupervised object discovery is becoming an essential line of research for tackling recognition problems that require decomposing an image into entities, such as semantic segmentation and object detection. Recently, object-centric methods that leverage self-supervision have gained popularity, due to their simplicity and adaptability to different settings and conditions. However, those methods do not exploit effective techniques already employed in modern self-supervised approaches.
In this work, we consider an object-centric approach in which DINO ViT features are reconstructed via a set of queried representations called slots. Based on that, we propose a masking scheme on input features that selectively disregards the background regions, inducing our model to focus more on salient objects during the reconstruction phase. Moreover, we extend the slot attention to a multi-query approach, allowing the model to learn multiple sets of slots, producing more stable masks.
During training, these multiple sets of slots are learned independently
while, at test time, these sets are merged through Hungarian matching to obtain the final slots. Our experimental results and ablations on the PASCAL-VOC 2012 dataset show the importance of each component and highlight how their combination consistently improves object localization. Our source code is available at: \href{https://github.com/rishavpramanik/maskedmultiqueryslot}{github.com/rishavpramanik/maskedmultiqueryslot}
\end{abstract}


\section{Introduction}
Modern-day computer vision technologies demand robust object localization methods for contemporary AI systems to understand and comprehend the ever-growing visual data. As our world transitions to become more dependent on AI-based systems, we observe that tasks like autonomous driving~\cite{wu2017squeezedet}, drone surveillance~\cite{xia2018dota}, and traffic monitoring~\cite{wang2017efficient} rely significantly on the generalization ability of object discovery tasks like object detection and segmentation. Current state-of-the-art approaches for object localization tasks heavily depend on meticulously labelled high-quality training data~\cite{lin2014microsoft,girshick2015fast}. Supervised approaches specialize in discovering a small subset of classes present in the physical world, thereby adversely impacting generalization. As an evolving subfield of representation learning, object-centric representation learning considers a visual scene to encompass multiple objects, potentially offering greater robustness against out-of-distribution data~\cite{dittadi2022generalization}. In this work, we consider unsupervised object discovery as object-centric representation learning.

Conventional approaches focusing on unsupervised localization of objects explore saliency~\cite{zitnick2014edge} or similarity within patches~\cite{uijlings2013selective} of an image. In contrast, other studies focus on utilizing data from an entire image collection by exploiting inter-image similarities via ranking~\cite{vo2021large}, component analysis~\cite{wei2019unsupervised}, and probabilistic matching~\cite{cho2015unsupervised}, but these methods hardly scale to large datasets. The latest methods use pre-trained self-supervised features~\cite{caron2021emerging,he2022masked} to localize objects. Generally, these methods~\cite{simeoni2021localizing,wang2022self} assume the presence of a single object that covers the majority of the image. However, such assumptions limit the application areas of these methods. The literature also includes works that use spectral clustering~\cite{zadaianchuk2022unsupervised,melas2022deep}; these methods typically necessitate an extra segmentation network to obtain the segmentation masks.

In DINOSAUR~\cite{seitzer2023bridging}, the authors propose to use an object-centric learning approach through slot attention~\cite{locatello2020object}. The authors use self-supervised features from DINO to reconstruct features with slot groupings. Although the authors highlight the usefulness of DINO features and slot attention groupings in decomposing a real-life visual scene into a collection of objects, the proposed approach does not fully exploit the potential of the DINO features and slot groupings. In this paper, we investigate the impact of masked modelling on DINO features in contributing to generalization towards more complex real-life images by leveraging object-centric representations. Instead of relying on single object groupings, we introduce additional inductive biases by producing multiple object groupings against a single write head to improve the stability and robustness of the framework.
During training, we feed non-overlapping patches to extract patch tokens from the last attention layer of a vision transformer (ViT) pre-trained with DINO~\cite{caron2021emerging}. We then selectively mask with zeros, $m\%$ of the patches with the highest means. These patches mostly belong to the background; thus, we leave the salient regions of the image unmasked. This allows the model to learn complex semantic information about the object. Next, we propose an extension of slot attention~\cite{locatello2020object}. During training, we consider the slots as queries and independently train multiple sets of slots against a single key-value pair. Then, we randomly select a single head of slot representations to be decoded. This strategy helps to produce independent representations at test time. During inference, we perform a fusion of the multiple heads by maximizing the cosine similarity among the slots to obtain the best matching amongst the learned representations. Finally, we obtain the aligned representations and compute an average across all the heads to produce the final set of output slots.

\textbf{Our main contributions} are as follows: (1) We introduce a strategy for selective masking of DINO features during training, which helps to better discriminate between different objects. (2) We propose to extend slot attention with multi-query attention along with different combinations training and inference techniques which help to stabilize the framework and enhance generalization. (3) We demonstrate the usefulness of each component of our approach for unsupervised object discovery, through different performance metrics.
\section{Related Work}
\subsection{Unsupervised Object Discovery}
Unsupervised object discovery aims to identify and group similar objects, subsequently localizing them within an image. Traditionally in literature, the initial task had been to find binary masks for object identification and grouping~\cite{li2015weighted,yan2013hierarchical}, while the subsequent task had involved outlining bounding boxes around these objects~\cite{jiang2013salient,kim2009unsupervised}.

It is not until very recently, that these distinct tasks have been tackled simultaneously by leveraging features from large-scale pre-trained methods. In this domain there have been some attempts that employ CNN activations for object discovery.~\cite{zhang2020object,collins2018deep}. LOST~\cite{simeoni2021localizing} was one of the early works which leveraged self-supervised trasnformer features for this task. The authors used a seed expansion strategy based on a graph constructed by using feature similarity extracted from self-supervised ViTs. Further extending this line of research, the work reported in Tokencut~\cite{wang2022self} and Deep spectral methods~\cite{melas2022deep} use normalized graph-cuts on this graph. These methodologies operate under the assumption that only one object is present in the image, modelling it as a bipartition problem in the graph, separating the background and foreground. Although these methods typically do not consider training due to their simplistic nature, FOUND~\cite{simeoni2023unsupervised} uses a $1\times 1$ convolution filter to refine coarse background masks to a more refined mask. DINOSAUR~\cite{seitzer2023bridging} is more closely related to our work and performs representation learning by reconstructing the self-supervised features extracted from the ViTs. Our work offers substantial improvement in performance and stability over DINOSAUR through selective masking strategies and multi-query attention.

\subsection{Self-Supervised learning}
The rapid advancement of self-supervised visual representation learning has opened ways to cut dependence on large-scale labelled datasets. Initially, such pretext training tasks were based on jigsaw puzzles, colorization and inpainting~\cite{doersch2015unsupervised,noroozi2016unsupervised,pathak2016context}. Recently, contrastive learning approaches which strongly rely on data augmentations are getting popular. These approaches require one or more image views, which are contrasted to learn feature representations~\cite{he2020momentum,misra2020self,chen2020simple}. Clustering-based approaches use discriminative training objectives between groups of images to learn feature representations~\cite{caron2020unsupervised,asano2020self}. In these approaches, a computed pseudo label acts as the cluster. All these techniques focus on downstream tasks such as image classification, emphasizing the extraction of global features. Self-supervised methods like DINO~\cite{caron2021emerging} enable the extraction of dense features for image patches, containing precise information about object presence in a patch.
\subsection{Masked Autoencoders}
Masked image modelling learns visual representations from corrupted or incomplete images, often deliberately performed. One of the early works~\cite{vincent2008extracting} that spearheaded research in this domain considers masking as noise for denoising autoencoders. Recently, frameworks like~\cite{bao2022beit,xie2022simmim} proposed using ViTs for masked language modelling. Masked Autoencoders (MAE)~\cite{he2022masked} is one of the recent advancements in self-supervised learning. MAEs randomly mask ViT patches at a relatively higher ratio than their language counterparts. To the best of our knowledge, there is no previous work which uses masked image modelling for object discovery without labels.
\subsection{Multi-Query Attention}
Neural attention is one of the most effective strategies in modern representation learning. This process involves comparing queries with keys to derive weights for the output, generally calculated as a weighted sum of value vectors. Multi-head attention introduced with the transformer model~\cite{vaswani2017attention} splits the keys, queries and values into multiple heads to derive the weights for the attention layer outputs. This idea was improved in further work by Shazeer~\cite{shazeer2019fast} which proposed the use of a single key and value pair for faster inference and reduced memory footprint. We exploit the idea of having multiple heads with a single key and value pair to improve on Slot Attention~\cite{locatello2020object}.

\section{Methodology}
\begin{figure*}[ht!]
	\centering  
		\includegraphics[width=0.92\textwidth]{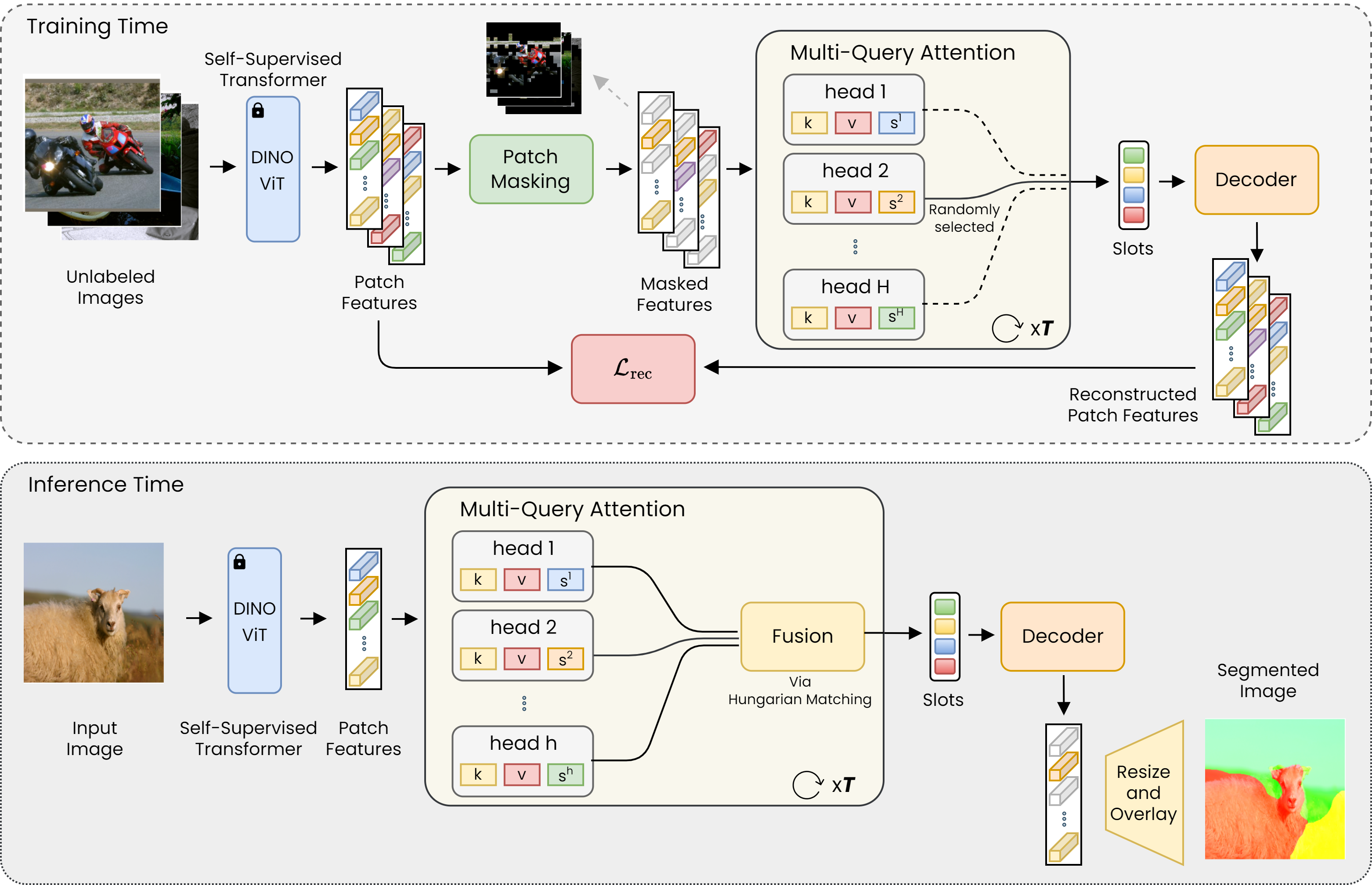} 
	  \caption[Overview of our architecture]{Overview of our proposed architecture. The top part describes the training pipeline that highlights two of our main contributions: Patch Masking Strategy and Multi-Query Attention. The bottom part illustrates the inference phase, which unlike the training phase, does not include the patch masking module and includes an additional Fusion module to perform Slot Alignment via Hungarian Matching to produce the output slots.
}\label{fig:architecture}
\end{figure*}
In this section, we discuss our proposed method in detail. We start by introducing the masking strategy and the multi-query slot attention approach. Then, in the subsequent sub-sections, we discuss the combination of the multiple queries. We also discuss about how we build the entire end-to-end framework for unsupervised object discovery. We present an overview of our method in Fig.~\ref{fig:architecture}.

\subsection{Masking}
\begin{figure*}
    \centering
    \includegraphics[width=0.16\linewidth,keepaspectratio]{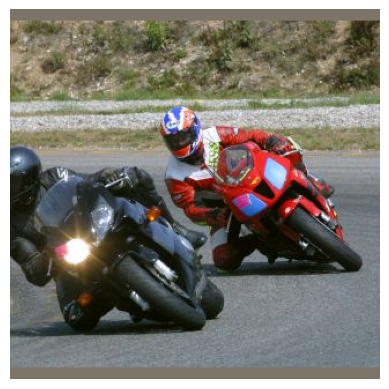}
    \includegraphics[width=0.16\linewidth,keepaspectratio]{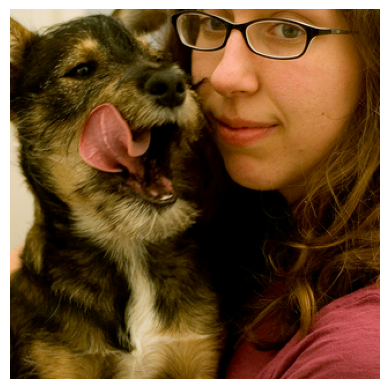}
    \includegraphics[width=0.16\linewidth,keepaspectratio]{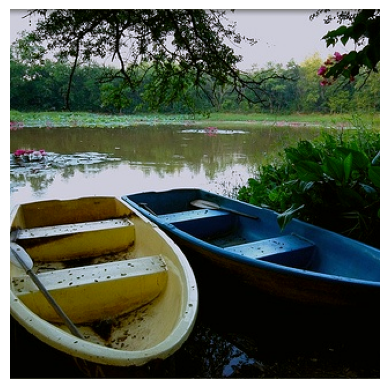}
    \includegraphics[width=0.16\linewidth,keepaspectratio]{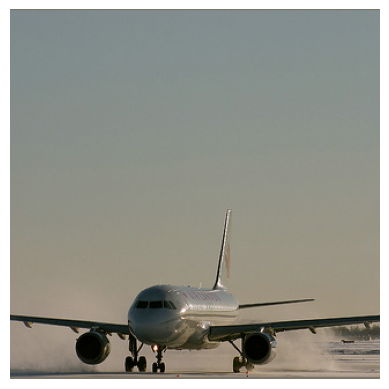}
    \includegraphics[width=0.16\linewidth,keepaspectratio]{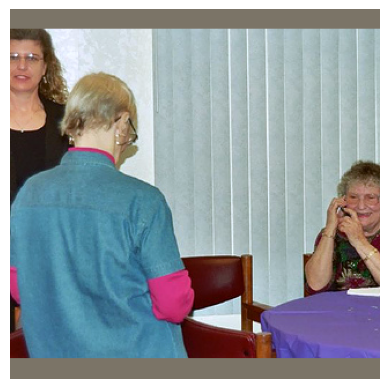}
    \includegraphics[width=0.16\linewidth,keepaspectratio]{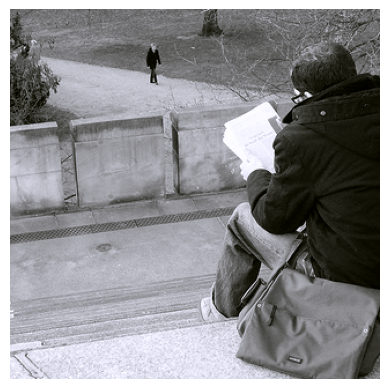}

    \includegraphics[width=0.16\linewidth,keepaspectratio]{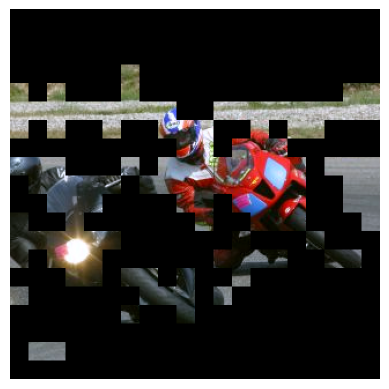}
    \includegraphics[width=0.16\linewidth,keepaspectratio]{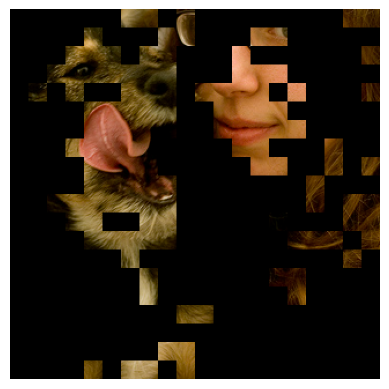}
    \includegraphics[width=0.16\linewidth,keepaspectratio]{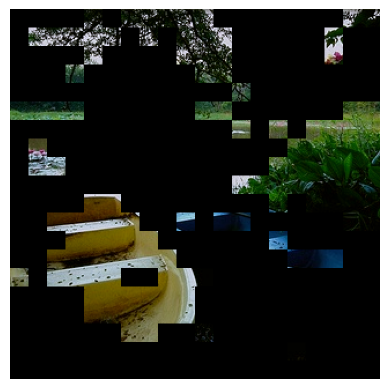}
    \includegraphics[width=0.16\linewidth,keepaspectratio]{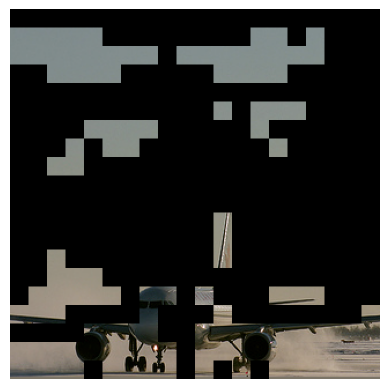}
    \includegraphics[width=0.16\linewidth,keepaspectratio]{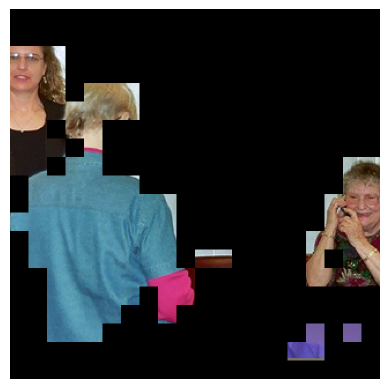}
    \includegraphics[width=0.16\linewidth,keepaspectratio]{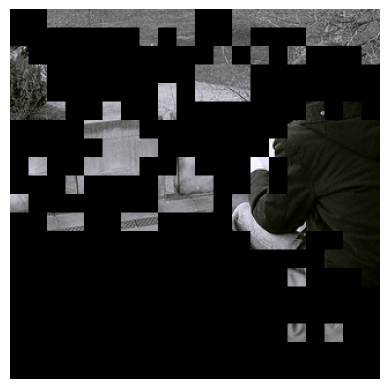}
    \caption{Masking visualization. These figures show the original image (first row) and the corresponding masked version used during training. As expected, masking removes a large portion of the background leaving only parts of the object of interest and forcing the model to reconstruct the missing parts.}
    \label{fig:masking}
\end{figure*}

We devise our masking strategy with inspiration from the MAEs literature \cite{he2022masked}. 
Let us consider an image $x$ with spatial resolution $H\times W$ pixels reshaped into $N=\frac{H\cdot W}{P^2}$ non-overlapping patches and with $P$ the patch size in pixels. Each patch token extracted from ViT-S/16 pre-trained with DINO is represented by a vector $p \in \mathbb{R}^{D_{\text{feats}}}$, with dimensionality $D_{\text{feats}}$. We compute the mean ($ \frac{1}{D_{\text{feats}}}\sum_{i=1}^{D_{\text{feats}}}p_i $
) for each patch token to obtain a rough estimate of the saliency of the patch w.r.t. to the entire image. We then sort the patches in increasing order and filter out the top $m\%$ of the patches with the highest means. 
We mask these patches by replacing the patch tokens with a zero vector $\vec{0}$. 
Implementation-wise, we use the $arg\,sort$ operator from standard libraries to obtain the indices in increasing order. We then filter out the last $m\%$ of the indices to mask them. Under this framework, we assume the objects usually have a more complex semantic structure when compared to their background counterparts. Thus, we leave the majority of the objects unmasked. This strategy permits the model to learn invariant object-centric representations, allowing us to generate generalizable semantic masks. As an added advantage of primarily masking the background, our method discriminates well between different sets of objects. We present a few examples of our masking process in Fig.~\ref{fig:masking}.
\subsection{Multi-Query Slot Attention}
Our multi-query slot attention approach is based on slot attention~\cite{locatello2020object}. Each slot $s$ in a slot attention module is a representation of dimension $(D_{\text{slots}})$ that serves to bind an object to different categories through an iterative dot-product attention process. An encoder network $(E)$ encodes a given image into features $(F)$ based on which slot attention mimics a competition process among $K$ slots (queries) initialized randomly using:
\begin{gather}
    \label{eqn:randomly}
    slots \sim \mathcal{N}(\mu,diag(\sigma)) \in \mathbb{R}^{K \times D_{\text{slots}}}
\end{gather}
Here $\mu$ and $\sigma$ are the learnable mean and variance of a Gaussian. The mean and variance are shared amongst the slots, \textit{i.e.}, we use the same mean and variance to initialize all the slots. In our formulation, we define $h$ heads of slots and we initialize these sets of slots $(s^j : j \in (1,2...h))$ with mean and variance $\mu^j$ and $\sigma^j$, respectively. We then use learnable linear projections $(K,V,Q^j)$ for attention. To learn the projections on keys $(k)$, values $(v)$ and queries (slots $s^j$) on inputs $x$ we use:
\begin{gather}
    \label{eqn:projection}
    k=K(x) \;,\; v=V(x) \;,\; s^j=Q^j(s^j)
\end{gather}
Our multi-query attention module uses only a single key-value pair; therefore, we apply a single set of transformations for all heads, as opposed to the queries for which we use different transformations for each head. We follow the strategy of Multi-Query Attention~\cite{shazeer2019fast}, which proposes using only one write head. We also consider that slot attention is an iterative process and hence, in each iteration, the projection is learned using the same transformation on the slots.

We compute the dot product attention for each head $j$ using:
\begin{gather}
    \label{eqn:attn}
    A^j=\frac{e^{\lambda^j}}{\sum_{\forall \lambda^j} e^{\lambda^j}} \;:\;  \lambda^j=(\frac{1}{\sqrt{D_{\text{slots}}}})k\cdot (s^j)^T \in \mathbb{R}^{K\times D_{\text{slots}}}\\
    \label{eqn:wattn}
    \upsilon^j=(\mathcal{W}^j)^T\cdot v \; :\; \mathcal{W}^j= \frac{A^j+\epsilon}{\sum_{\forall A^j}(A^j+\epsilon)}
\end{gather}
In our approach, we evaluate independent attention weights for each head. These weights help us focus on relevant regions in the image against the set of queries (slots) that we initialize with mean and variance for a particular head. By introducing this mechanism of having independent heads, we average out the negative impact of seed selection over different regions, which we observe as a problem in various previous works~\cite{simeoni2021localizing,wang2022self,seitzer2023bridging}. This idea of having different slot proposals improves the stability of the entire framework. The slot representations in each head are learned through an iterative process with $t \in (1,2...T)$ iterations. We update the slots $s_t^j$ at the $t^{th}$ iteration using a Gated Recurrent Unit (GRU) and multi-layer perceptron (MLP) using: 
\begin{gather}
    \label{eqn:updategru}
    s_t^j\mathrel{+}=\text{MLP}^j(s_t^j) : s^j_t=\text{GRU}^j(\text{state}=s_{t-1}^j,\text{input}=\upsilon^j)
\end{gather}
\subsection{Combination of Multiple Queries}
The multi-query attention process demands a mechanism to aggregate the slot representations across various heads. In this regard, we consider a crucial property of Slot Attention: it maintains permutation equivariance concerning the order of the slots~\cite{locatello2020object}. This property implies that rearranging the slot order after initialization equates to rearranging the output order of the module. Another key factor that we take into account while combining the slots is the computational cost, as well as, the time required for both training and inference.

\subsubsection{Training}
During the training phase, we prioritize the ability of each head to offer independent slot representations. To accomplish this, we ensure that all parameters related to a particular head—such as the means, variance, MLP projections, and GRU—are trained independently. During batch training at runtime, we randomly select a head to be processed by the decoder. This random selection enables each parameter associated with each head to learn distinctive features while maintaining a low computational cost.
\begin{figure*}
    \centering
    \includegraphics[width=0.168\linewidth,keepaspectratio]{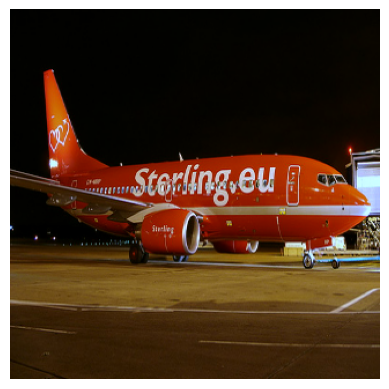}
    \includegraphics[width=0.168\linewidth,keepaspectratio]{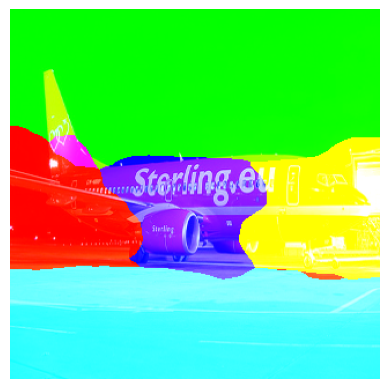}
    \includegraphics[width=0.168\linewidth,keepaspectratio]{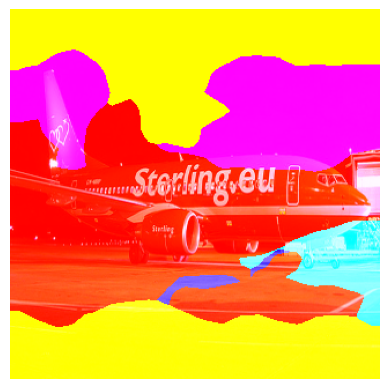}
    \includegraphics[width=0.168\linewidth,keepaspectratio]{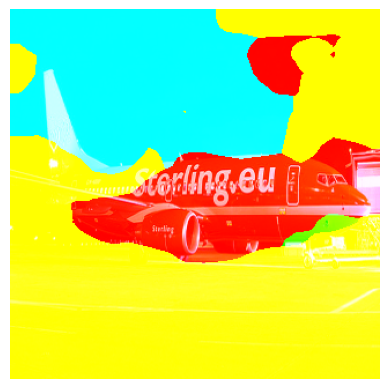}

    \includegraphics[width=0.168\linewidth,keepaspectratio]{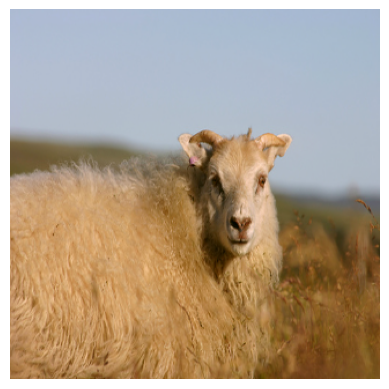}
    \includegraphics[width=0.168\linewidth,keepaspectratio]{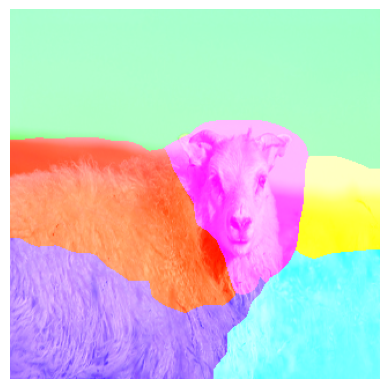}
    \includegraphics[width=0.168\linewidth,keepaspectratio]{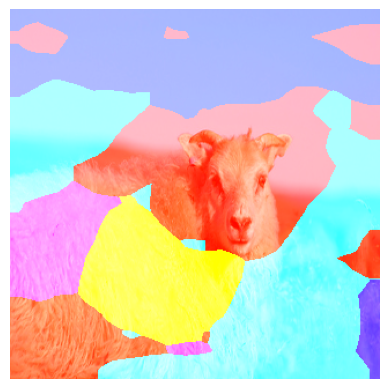}
    \includegraphics[width=0.168\linewidth,keepaspectratio]{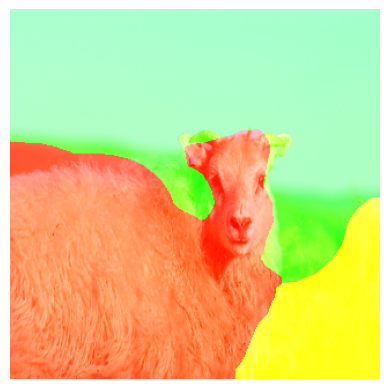}

    \includegraphics[width=0.168\linewidth,keepaspectratio]{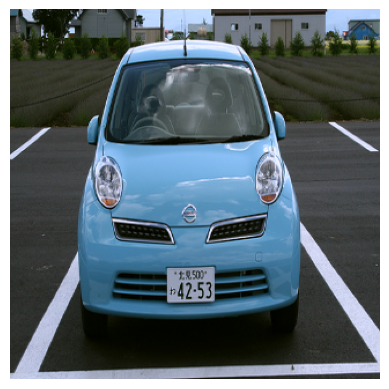}
    \includegraphics[width=0.168\linewidth,keepaspectratio]{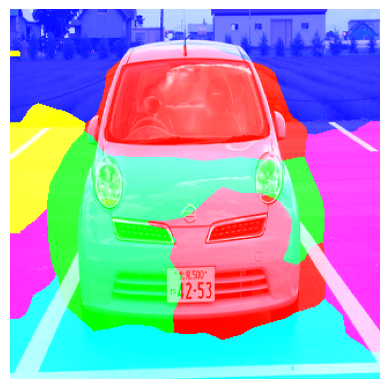}
    \includegraphics[width=0.168\linewidth,keepaspectratio]{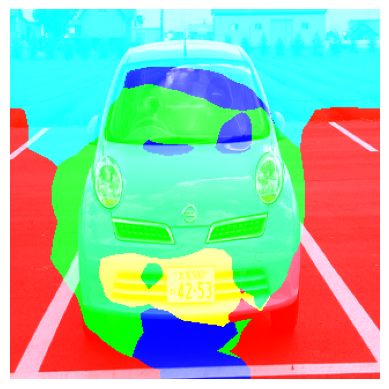}
    \includegraphics[width=0.168\linewidth,keepaspectratio]{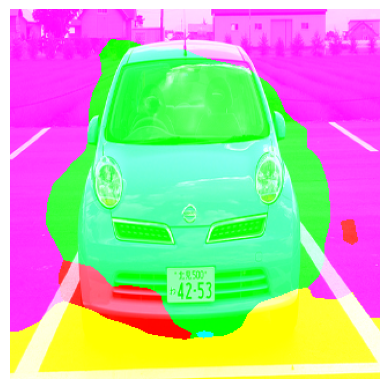}

    \includegraphics[width=0.168\linewidth,keepaspectratio]{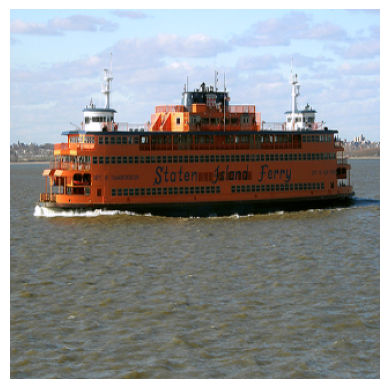}
    \includegraphics[width=0.168\linewidth,keepaspectratio]{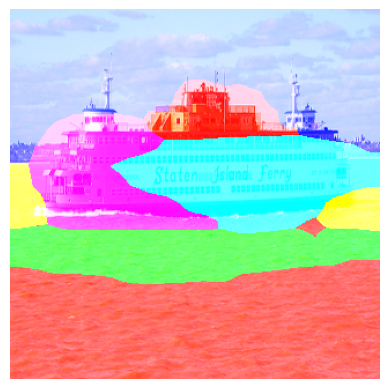}
    \includegraphics[width=0.168\linewidth,keepaspectratio]{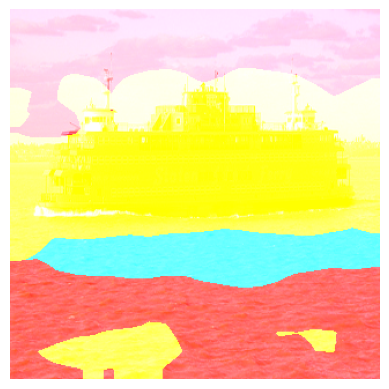}
    \includegraphics[width=0.168\linewidth,keepaspectratio]{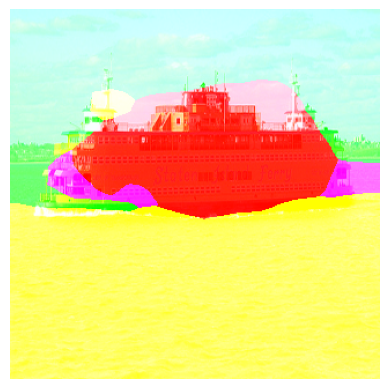}
    \caption{Segmentation masks created with different slot attention configurations. From left to right: Original image, normal Slot Attention, Slot Attention with our masking approach, Slot Attention with masking and Multi-Queries (Ours). All configurations involve a ViT-S/16 pre-trained with DINO as the encoder backbone.}
    \label{fig:compare}
\end{figure*}
\subsubsection{Inference}
Since we know, that changing the slot order is sufficient to align the two given slots, let us consider a set of slot representations $(s^j)$ for head $j \in (1,2...h)$. We randomly select a reference head $(s^r)$ within the $h$ heads, based on which we align the remaining sets of slots. We consider a single slot representation from $s^j$ as $c^j_i$ for the slot $i \in (1,2...K)$. Then, to align the slot representations, we obtain the similarity matrix using cosine similarity according to:
\begin{gather}
    \label{eqn:cos}
    S_{jr}=\frac{c^j_a \cdot c^r_b}{||c^j_a||_2 ||c^r_b||_2}\; \forall\; a,b 
\end{gather}
We match the respective slots from head $j$ to the reference head $r$ using Hungarian Matching~\cite{kuhn1955hungarian}. We maximize the cosine similarity to obtain the matching. To obtain the final set of slot representations $(s^f)$ that will be decoded we use:
\begin{gather}
    s^f=\frac{1}{h}(s^r+\sum_{\forall s^j} \text{aligned} (s^j)) \; : \; j\neq r 
\end{gather} 
\textit{i.e.} the averaged representations corresponding to the aligned slots.
In this work, we also compare a few other combination criteria, which are discussed in the \hyperref[sec:res]{Results} section.
\subsection{Architecture}
\subsubsection{Encoding}
The encoder extracts feature representations from the ViT-S/16 model trained in a self-supervised fashion using DINO~\cite{caron2021emerging}. It learns three linear projections—\textbf{K}eys, \textbf{Q}ueries, and \textbf{V}alues—based on non-overlapping patches from the input image. Of the multiple self-attention layers, we extract the patch tokens from the last self-attention layer, typically composed of many heads. In this paper, we concatenate all keys from these heads. Although we refrain from fine-tuning the ViT in this paper, our efforts to do so were computationally impractical.

\subsubsection{Decoding}
We use a decoder to transform each slot representation into patch tokens. We use the spatial broadcast decoder~\cite{watters2019spatial}, a common decoding strategy~\cite{locatello2020object,seitzer2023bridging} for the latent vectors in variational autoencoders. Initially, we broadcast a given slot into the number of patches. We then add additive positional encodings to the slot representations similar to the positional encodings in ViT. We further process these features through a 3-layer MLP with a hidden depth of $1024$ to obtain the patch tokens for each slot. We follow a process similar to previous works~\cite{locatello2020object,greff2019multi} to normalize the patch tokens across the slots using the $softmax$ operation. Finally, we use these weights to produce the reconstructed patch tokens as a weighted sum of all the decoded slot representations. We use a mean squared error ($\mathcal{L}_2$) loss to optimize our network.
\section{Results}
\label{sec:res}
In this section, we present our experimental results and their corresponding analysis. We begin by discussing the importance of our masking strategy compared to other methods. We provide results related to different multi-head combination strategies. Following that, we conduct an ablation study to explore the impact of various aspects of our approach. In the latter part of this section, we present the associated computational cost. Finally, we compare our method to the current state-of-the-art unsupervised object discovery methods to conclude our results. All our experiments are conducted on the PASCAL-VOC 2012 dataset~\cite{everingham2015pascal}.

\subsection{Preliminaries}
We enlist some important implementation details and the metrics we use to evaluate our method as follows:
\subsubsection{Implementation Details}
We train the masked multi-query slot attention using the Adam optimizer. We use a learning rate of $4\times 10^{-4}$, and we set $2\%$ of the total optimization steps to linearly warm up the optimization process. We then use an exponential decay to allow our method to converge. We train for $500$ epochs on $4\times$ $16$ GB NVIDIA-V100 GPUs with a local batch size of $4$. For reporting each result, we run each method $3$ times and report the mean and standard deviation. During training for the reconstruction, the probability of masking a patch is $m=70\%$, which is determined empirically.
\subsubsection{Evaluation Metrics}
To fairly assess our experimental findings, we report Correct Localization (CorLoc), mean intersection over union (mIoU), and mean best overlap (mBo). In the CorLoc metric, we gauge the percentage of correct bounding boxes, selecting the predicted boxes with IoU greater than 0.5 with any of the ground truth boxes. We calculate mIoU and mBo by comparing the class-agnostic segmentation masks to the ground truth masks. For this task, we perform Hungarian matching by maximizing the IoUs to obtain the matching. The mIoU measures the IoU per class and averages it over all the classes per image for the entire dataset. The mBo is the average IoU over the assigned mask pairs.
\subsection{Baselines}

In Table~\ref{tab:overall} and Fig.~\ref{fig:compare}, we compare our method against two other baselines. For each method, we use ViT-S/16 pre-trained with DINO as the feature extractor. The masking strategy we use for comparison in Table~\ref{tab:overall} and Fig.~\ref{fig:compare} is the background masking. We show that our masking produces substantially better results both quantitatively and qualitatively. Through masking background (Fig.~\ref{fig:masking}), we ensure our model learns to group the objects within an image in the absence of background noise.
Finally, by introducing the multi-query approach, we further boost performance.
\begin{table}[ht!]
    \caption{Comparative results with different masking and multi-query configurations. All configurations use a ViT-S/16 pre-trained with DINO as the backbone network}
    \begin{adjustbox}{max width=\linewidth}
    \begin{tabular}{cc|ccc}
    \toprule
    Masking & Multi-Query & CorLoc & mIou\textsuperscript{i} & mBo\textsuperscript{i}\\
    \midrule
    \ding{56} & \ding{56} & 50.38 $\pm$ 0.9 & 32.82 $\pm$ 0.1 & 32.92 $\pm$ 0.1\\
    \ding{51} & \ding{56} &  64.48 $\pm$ 0.4 &  36.35 $\pm$ 1.3 & 37.34 $\pm$ 0.4\\
    \ding{51} & \ding{51} & 68.99 $\pm$ 0.7& 39.42 $\pm$ 0.3 & 39.74 $\pm$ 0.3\\
    \bottomrule
    \end{tabular}
    \end{adjustbox}
    \label{tab:overall}
\end{table}
\subsection{Masking}

We also compare our masking approach against the commonly used~\cite{he2022masked} random masking strategy. We randomly select the patch indices in the runtime and replace the patch tokens with zeros. In Table~\ref{tab:maskings}, we observe that background masking produces better and more stable results for all three metrics used for comparison here. The improvement can be explained noting that, unlike background masking, random masking masks patch features for both the object and the background, thus not allowing the model to learn semantic information about a specific object. While background masking is a bit more expensive than random masking in the training phase, the inference time remains the same for both.
\begin{table}[ht!]
    \centering
    \caption{Results with different masking strategies. \ding{51} indicates if multiple heads was used during both training and evaluation}
    \begin{adjustbox}{max width=\linewidth}
    \begin{tabular}{cc|ccc}
    \toprule
    Strategy & Multiple Heads & CorLoc & mIou\textsuperscript{i} & mBo\textsuperscript{i}\\
    \midrule
    \multirow{2}{*}{No Masking} & \ding{56} & 50.38 $\pm$ 0.9 & 32.82 $\pm$ 0.1 & 32.92 $\pm$ 0.1\\
    & \ding{51} & 51.22 $\pm$ 0.8  &33.06 $\pm$ 0.1 & 30.96 $\pm$ 0.2 \\
    \midrule
    \multirow{2}{*}{Random} & \ding{56} & 67.35 $\pm$ 6.2 & 34.60 $\pm$ 2.5 & 34.68 $\pm$ 2.4\\
    & \ding{51} & 66.60 $\pm$ 1.7 & 32.64 $\pm$ 0.1 & 32.79 $\pm$ 0.1\\
    \midrule
    \multirow{2}{*}{Background} & \ding{56} & 64.48 $\pm$ 1.4 & 36.35 $\pm$ 0.4 & 37.34 $\pm$ 0.4\\
    & \ding{51}  & 68.99 $\pm$ 0.7& 39.42 $\pm$ 0.3 & 39.74 $\pm$ 0.3\\
    \bottomrule
    \end{tabular}
    \end{adjustbox}
    \label{tab:maskings}
\end{table}
\begin{figure*}[ht!]
    \centering
    \includegraphics[width=0.088\linewidth,keepaspectratio]{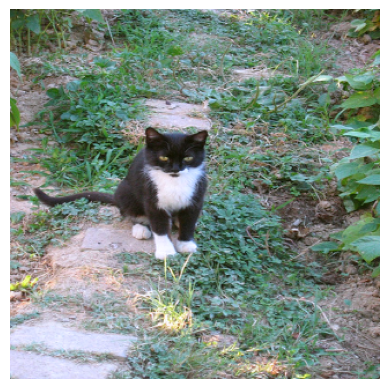}
    \includegraphics[width=0.088\linewidth,keepaspectratio]{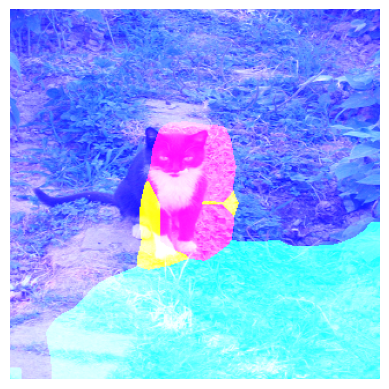}
    \includegraphics[width=0.088\linewidth,keepaspectratio]{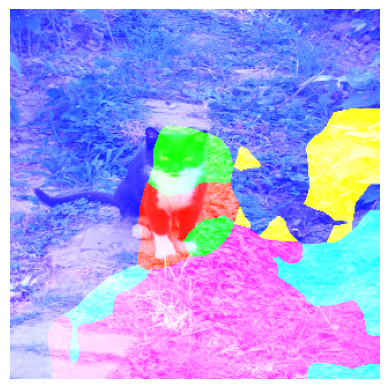}
    \includegraphics[width=0.088\linewidth,keepaspectratio]{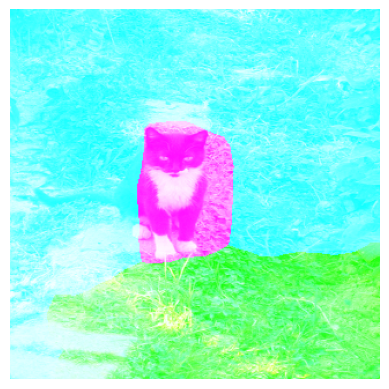}
    \includegraphics[width=0.088\linewidth,keepaspectratio]{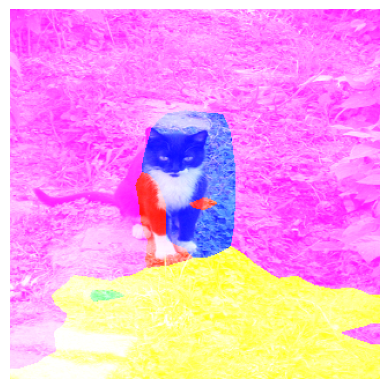}
    \includegraphics[width=0.088\linewidth,keepaspectratio]{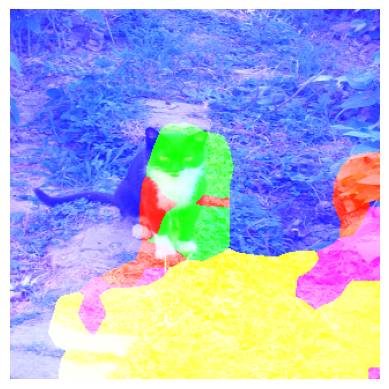}
    \includegraphics[width=0.088\linewidth,keepaspectratio]{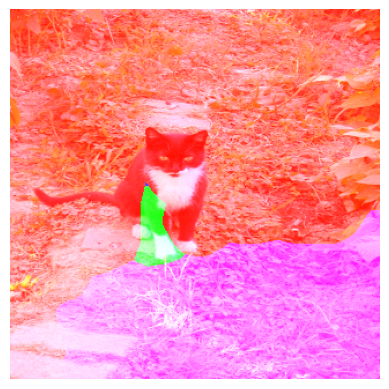}
    \includegraphics[width=0.088\linewidth,keepaspectratio]{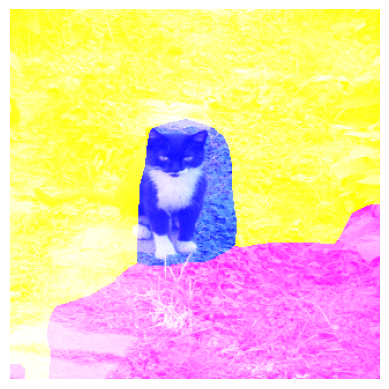}
    \includegraphics[width=0.088\linewidth,keepaspectratio]{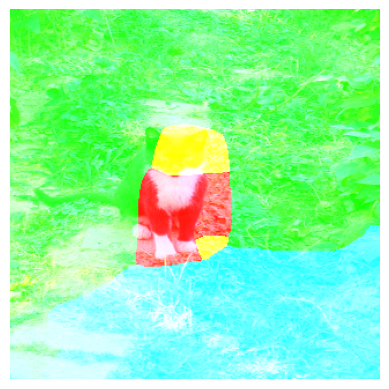}
    \includegraphics[width=0.088\linewidth,keepaspectratio]{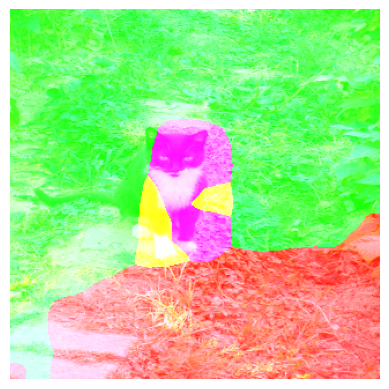}

    \includegraphics[width=0.088\linewidth,keepaspectratio]{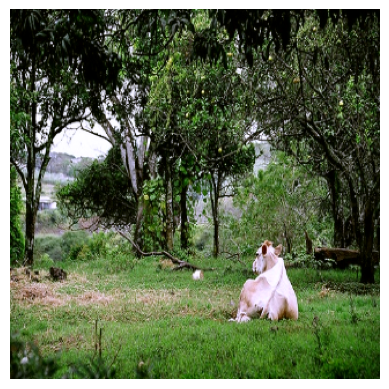}
    \includegraphics[width=0.088\linewidth,keepaspectratio]{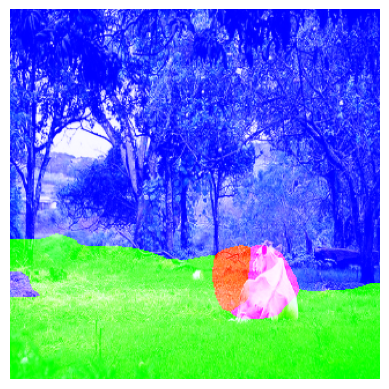}
    \includegraphics[width=0.088\linewidth,keepaspectratio]{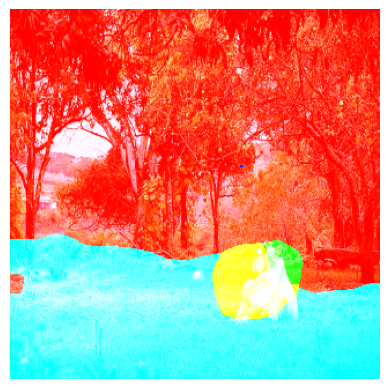}
    \includegraphics[width=0.088\linewidth,keepaspectratio]{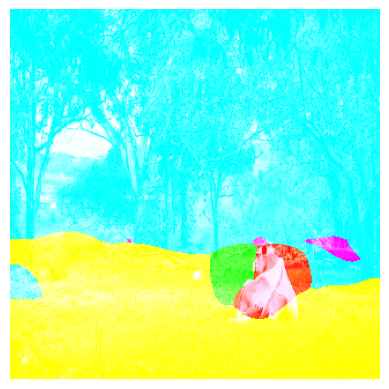}
    \includegraphics[width=0.088\linewidth,keepaspectratio]{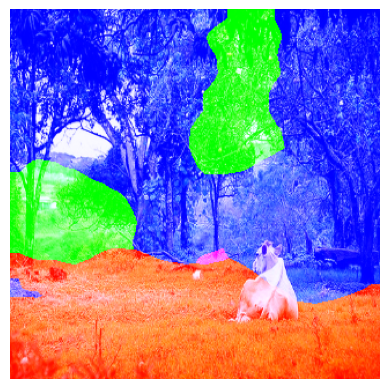}
    \includegraphics[width=0.088\linewidth,keepaspectratio]{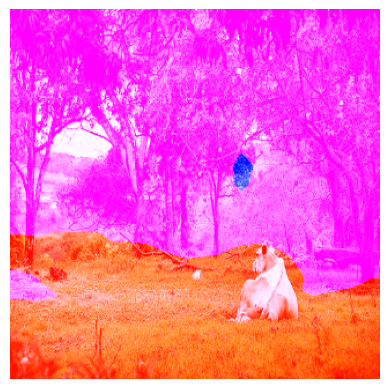}
    \includegraphics[width=0.088\linewidth,keepaspectratio]{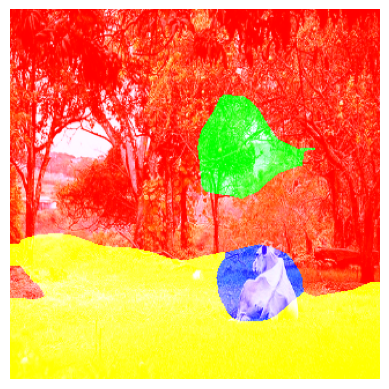}
    \includegraphics[width=0.088\linewidth,keepaspectratio]{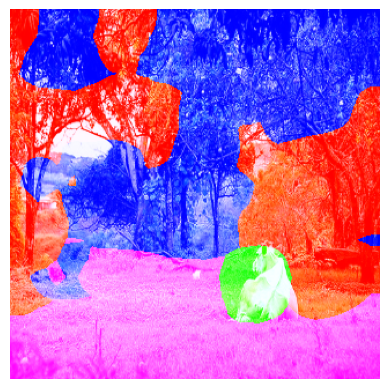}
    \includegraphics[width=0.088\linewidth,keepaspectratio]{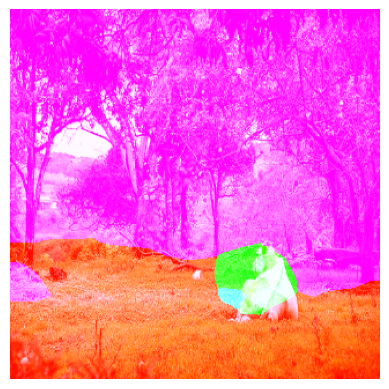}
    \includegraphics[width=0.088\linewidth,keepaspectratio]{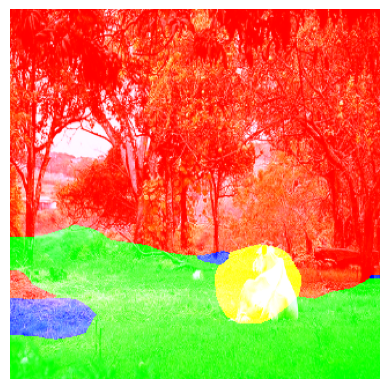}
    
    \caption{Segmentation masks of the different heads for the multi-query approach. The first column on the left is the original Image. The next 8 images show the different segmentations for each head and the rightmost column shows the combined results with Hungarian matching.}
    \label{fig:head}
\end{figure*}
\subsection{Multi Query Attention and Combination}
We show the visualizations in Fig.~\ref{fig:head} concerning each head and the corresponding visualization post combination through Hungarian matching. We observe that a few heads produce noisy masks and note that our strategy of training with independent heads helps reduce this bias with other heads helping to stabilize the final output.
\begin{table}[ht!]
    \centering
    \caption{Analysis of different combination schemes for multi-query attention. Note that all models involve background masking as the masking strategy}
    \begin{adjustbox}{max width=\linewidth}
    \begin{tabular}{ccc |ccc}
    \toprule
    Distance & Matching & Training & CorLoc & mIou\textsuperscript{i} & mBo\textsuperscript{i}\\
    \midrule
    \multirow{4}{*}{Euclidean} & \multirow{2}{*}{Greedy} & \ding{56} & 58.62 $\pm$ 1.0 & 27.84 $\pm$ 0.9 & 21.88 $\pm$ 1.1\\
     & & \ding{51} & 50.75 $\pm$ 2.3 & 33.39 $\pm$ 0.8 & 23.08 $\pm$ 1.4\\
     \cmidrule{2-6}
     & \multirow{2}{*}{Hungarian} & \ding{56} & 67.03 $\pm$ 1.8 & 38.1 $\pm$ 0.6 & 29.89 $\pm$ 1.8\\
     & & \ding{51} & 58.04 $\pm$ 0.7 &  28.64 $\pm$ 0.3& 23.68 $\pm$ 0.6 \\
     \midrule
    \multirow{4}{*}{Cosine} & \multirow{2}{*}{Greedy} & \ding{56} & 60.50 $\pm$ 0.6 & 29.93 $\pm$ 1.0 & 23.59 $\pm$ 1.1\\
     & & \ding{51} & 43.80 $\pm$ 1.4 & 31.73 $\pm$ 0.3 & 20.13 $\pm$ 0.5\\
     \cmidrule{2-6}
     & \multirow{2}{*}{Hungarian} & \ding{56} & 68.99 $\pm$ 0.7& 39.42 $\pm$ 0.3 & 39.74 $\pm$ 0.3 \\
     & & \ding{51} & 61.44 $\pm$ 3.8& 32.26 $\pm$ 1.4 & 32.49 $\pm$ 1.3\\
     
    \bottomrule
    \end{tabular}
    \end{adjustbox}
    \label{tab:head}
\end{table}
We compare various head combination strategies in Table~\ref{tab:head}. In our analysis, we compare Euclidean and Cosine distance metrics. We use this distance to compute the similarity amongst different pairs of heads to align them. Additionally, we employ a Hungarian or Greedy matching strategy to pair each slot. In the greedy strategy, after computing the similarity matrix, we perform the matching greedily. We iterate over the reference head and identify the best-matching slot from the head under consideration based on the similarity matrix.

We also compare based on training; the checkmark in the table indicates whether we use this matching strategy during training. The \ding{56} mark illustrates that we select a random head during training, and the combination is performed only during the inference phase. We observe that if we combine heads during the training phase, the performance of the model significantly drops. This drop indicates that each head does not produce independent object representations that stabilize the negative impact produced by other heads. We also observe that using the Hungarian matching strategy improves the result compared to other combination strategies.
\subsection{Additional Ablations}
\begin{figure}[ht!]
    \centering
    \includegraphics[width=0.45\linewidth,,keepaspectratio]{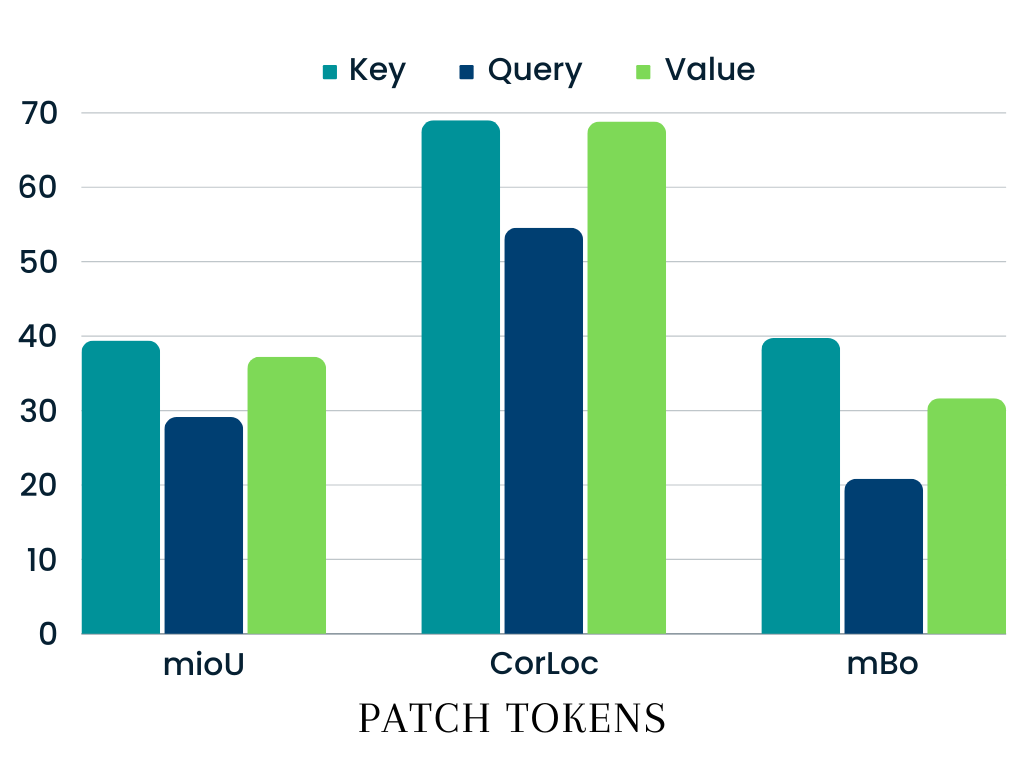}
    \caption{Ablation study concerning the use of different patch tokens namely key, query and value.}
    \label{fig:token}
\end{figure}

We present an ablation study on the number of heads in Table~\ref{tab:numheads} and the use of different patch tokens in Fig.~\ref{fig:token}. In Table~\ref{tab:numheads} we observe that by increasing number of heads the performance of the model increases and converges to a certain value. Notably, we also observe that the standard deviation also decreases as we increase the number of heads.
\begin{table}[ht!]
    \centering
    \caption{Ablation study concerning the number of heads in multi-query slot attention}
    \begin{adjustbox}{max width=\linewidth}
    \begin{tabular}{c|ccc}
    \toprule
    \#Heads & CorLoc & mIou\textsuperscript{i} & mBo\textsuperscript{i}\\
    \midrule
    1 & 64.48 $\pm$ 0.4 &  36.35 $\pm$ 1.3 & 37.34 $\pm$ 0.4\\
    2 & 67.40 $\pm$ 2.0 & 38.19 $\pm$ 0.4 & 37.54 $\pm$ 1.3\\
    4 &  69.79 $\pm$ 1.2 &  39.14 $\pm$ 0.4 & 39.46 $\pm$ 0.3\\
    8 & 68.99 $\pm$ 0.7& 39.42 $\pm$ 0.3 & 39.74 $\pm$ 0.3\\
    16 & 68.21 $\pm$ 1.2& 39.46 $\pm$ 0.1 & 39.82 $\pm$ 0.04\\
    \bottomrule
    \end{tabular}
    \end{adjustbox}
    \label{tab:numheads}
\end{table}
In Fig.~\ref{fig:token} we observe the performance of our model against using different patch tokens. We clearly notice that key tokens give slightly better results than the value tokens and that the query tokens produce sub-optimal results when compared to the other tokens.

\subsection{Computational Cost}
We discuss the computational cost associated with our model in Table~\ref{tab:params}. We compare total and trainable parameters, floating point operations per second (FLOPs), and the time required for forward and backward passes during training and inference.
\begin{table}[ht!]
    \centering
    \caption{Computational cost analysis with different configurations for a batch size of 4. We report the floating point operations per second (FLOPs) in millions, the number of parameters in millions and time required in milliseconds}
    \begin{adjustbox}{max width=\linewidth}
    \begin{tabular}{c|c|cc|ccc}
    \toprule
    \multirow{2}{*}{Configuration} & \multirow{2}{*}{FLOPs (M)} & \multicolumn{2}{c|}{Parameters (M)} & \multicolumn{3}{c}{Time (in ms)}\\
    \cmidrule{3-7}
    & & Total & Trainable & Forward & Backward & Inference\\
    \midrule
    Slot Attention & 151.16 & 26.82 & 5.2 & 17.69 & 24.92 & 20.02\\
    Maksed Slot Attention & 151.16 & 26.82 & 5.2 & 18.18 & 28.85 & 20.02\\
    Muti Query Attention & 151.23 & 36.84 & 15.2 & 30.36 & 38.20 & 55.72\\
    \bottomrule
    \end{tabular}
    \end{adjustbox}
    \label{tab:params}
\end{table}
In Table~\ref{tab:params}, we observe an increase in training time for models involving masking, which we attribute to evaluating each patch token to decide whether masking is required. However, it is noteworthy that the time required remains constant because masking is not involved in the inference phase. Finally, the inference time for our multi-query slot attention increases by a significant factor. However, our implementation is not optimized for speed. With a better implementation, we expect the cost of multi-query slot attention to become negligible.
Our proposed multi-query attention and combination involves a slight increase in computational cost, while the improvement in the result is quite significant.
\subsection{Comparison with previous methods}
We compare our method with several previous works in unsupervised object discovery. For this task, we compare our method against methods ranging from pre-deep learning era~\cite{uijlings2013selective,zitnick2014edge,zhang2020object} to CNN-based localization~\cite{zhang2020object,wei2019unsupervised} and graph-based methods~\cite{vo2021large}. We also compare our method with self-supervised ViT-based works~\cite{simeoni2021localizing,melas2022deep,wang2022freesolo,wang2022self,seitzer2023bridging}.
\begin{table}[ht!]
    \centering
    \caption{Comparative performance of our method against existing works in unsupervised object discovery}
    \begin{adjustbox}{max width=\linewidth}
    \begin{tabular}{c|c}
    \toprule
    Method & CorLoc \\
    \midrule
    Selective Search~\cite{uijlings2013selective} & 20.9\\
    Edge Boxes~\cite{zitnick2014edge} & 31.6\\
    Kim et al.~\cite{kim2009unsupervised} & 46.4\\
    Zhang et al.~\cite{zhang2020object} & 50.5\\
    Wei et al.~\cite{wei2019unsupervised} & 53.1\\
    rOSD~\cite{vo2021large} & 55.3\\
    DINO-seg (ViT-S/16)~\cite{caron2021emerging,simeoni2021localizing} & 46.2\\
    LOST (ViT-S/8)~\cite{simeoni2021localizing} & 57.0\\
    LOST (ViT-S/16)~\cite{simeoni2021localizing} & 64.0\\
    Deep Spectral Methods~\cite{melas2022deep} & 66.4\\
    TokenCut~\cite{wang2022self} & 72.1\\
    FreeSOLO~\cite{wang2022freesolo} & 56.7\\
    DINOSAUR~\cite{seitzer2023bridging} & 69.8 $\pm$ 4.9\\
    \midrule
    Ours & 68.99 $\pm$ 0.7\\
    \bottomrule
    \end{tabular}
    \end{adjustbox}
    \label{tab:sotacorloc}
\end{table}
We enlist the quantitative results in Table~\ref{tab:sotacorloc}. Compared to methods from the pre-deep learning period, the performance of our approach is significantly improved. CNN-based methods~\cite{zhang2020object,wei2019unsupervised} perform relatively well, but self-supervised ViT-based methods perform much better. One possible explanation for this can be traced to the fact that ViT implicitly induces an additional structural bias of decomposing each image into non-overlapping patches, which helps to evaluate each patch representation and correlate better with neighboring regions to localize an object. Among modern ViT approaches, only Tokencut~\cite{wang2022self} and DINOSAUR~\cite{seitzer2023bridging} obtain better results than our method. Most methods have an inclination to use a larger ViT model or a smaller patch size, while our experiments used ViT-Small with a patch size of 16.
\section{Conclusion}
In this paper, we presented some strategies to improve object discovery without labels in an object-centric representation learning approach. Contrary to the previous approaches, we mask the background during training, forcing the model to learn discriminative object representations by reconstructing the background. We then use multiple queries (slots), trained independently and combined during inference to produce more stable masks. Our detailed investigation with different settings showcases the strength and usability of each of the modules.

One obvious limitation of our method is the fixed number of slots. In future work, it would be interesting to investigate how models can dynamically select the number of slots based on an input image. In our analysis, we find how different masking strategies influence the model. Hence, we wish to investigate how we control the fine-grained knowledge that we introduce through these masking strategies. Another related line of work that we look forward to exploring is the robustness of our method towards out-of-distribution data and its ability to handle domain shifts.

\section*{Acknowledgment}
This research was supported by the National Science and
Engineering Research Council of Canada (NSERC), via its
Discovery Grant program and MITACS through the Globalink program. The authors acknowledge support
provided in part by Calcul Quebéc and the Digital Research Alliance of Canada.
\bibliographystyle{IEEEtran}
\bibliography{IEEEabrv,references}
\end{document}